\documentclass[10pt,twocolumn,letterpaper]{article}

\usepackage{iccv}
\usepackage{times}
\usepackage{epsfig}
\usepackage{graphicx}
\usepackage{amsmath}
\usepackage{amssymb}
\usepackage{bm}
\usepackage{colortbl}
\usepackage{graphics}
\usepackage{multirow}
\usepackage{tabularx}
\usepackage{tabu}

\usepackage{enumitem}
\usepackage{color}

\definecolor{bed}{RGB}{17, 128, 128}
\definecolor{books}{RGB}{247, 53, 58}
\definecolor{ceiling}{RGB}{101, 27, 201}
\definecolor{chair}{RGB}{52, 62, 246}
\definecolor{floor}{RGB}{230, 230, 237}
\definecolor{furniture}{RGB}{252, 71, 38}
\definecolor{objects}{RGB}{252, 32, 128}
\definecolor{painting}{RGB}{51, 54, 148}
\definecolor{sofa}{RGB}{221, 180, 143}
\definecolor{table}{RGB}{65, 248, 68}
\definecolor{tv}{RGB}{254, 213, 48}
\definecolor{wall}{RGB}{150, 150, 150}
\definecolor{window}{RGB}{45, 255, 254}

\providecommand{\fref}[1]{Figure~\ref{#1}}
\providecommand{\sref}[1]{Section~\ref{#1}}
\providecommand{\tref}[1]{Table~\ref{#1}}
\providecommand{\eg}{\emph{e.g.}}
\providecommand{\ie}{\emph{i.e.}}

\usepackage[breaklinks=true,bookmarks=false]{hyperref}

\iccvfinalcopy % *** Uncomment this line for the final submission

\def\iccvPaperID{4851} % *** Enter the ICCV Paper ID here

\newcommand{\bhline}[1]{\noalign{\hrule height #1}}

\ificcvfinal\fi
\begin{document}

%%%%%%%%% TITLE
\title{Incremental Class Discovery for Semantic Segmentation with RGBD Sensing}
%\title{Semantic- and Geometric-guided Incremental Clustering on Dense SLAM}

\author{Yoshikatsu Nakajima$^{1,2}$\\
\and
Byeongkeun Kang$^1$\\
\and
Hideo Saito$^2$\\
\and
Kris Kitani$^1$\\
\and
{${}^{1}$Carnegie Mellon University}\\
{\tt\small\{byeongkk,kkitani\}@andrew.cmu.edu}
\and
{${}^{2}$Keio University}\\
{\tt\small\{nakajima,saito\}@hvrl.ics.keio.ac.jp}
}

\maketitle
%\thispagestyle{empty}

%%%%%%%%% ABSTRACT
\begin{abstract}
This work addresses the task of open world semantic segmentation using RGBD sensing to discover new semantic classes over time. Although there are many types of objects in the real-word, current semantic segmentation methods make a closed world assumption and are trained only to segment a limited number of object classes. Towards a more open world approach, we propose a novel method that incrementally learns new classes for image segmentation. The proposed system first segments each RGBD frame using both color and geometric information, and then aggregates that information to build a single segmented dense 3D map of the environment. The segmented 3D map representation is a key component of our approach as it is used to discover new object classes by identifying coherent regions in the 3D map that have no semantic label.
The use of coherent region in the 3D map as a primitive element, rather than traditional elements such as surfels or voxels, also significantly reduces the computational complexity and memory use of our method. It thus leads to semi-real-time performance at {10.7}Hz when incrementally updating the dense 3D map at every frame. Through experiments on the NYUDv2 dataset, we demonstrate that the proposed method is able to correctly cluster objects of both known and unseen classes. We also show the quantitative comparison with the state-of-the-art supervised methods, the processing time of each step, and the influences of each component.

\end{abstract}

%%%%%%%%% BODY TEXT

\section{Introduction}

\begin{figure}[t]
\begin{center}
    \includegraphics[width=0.9\hsize]{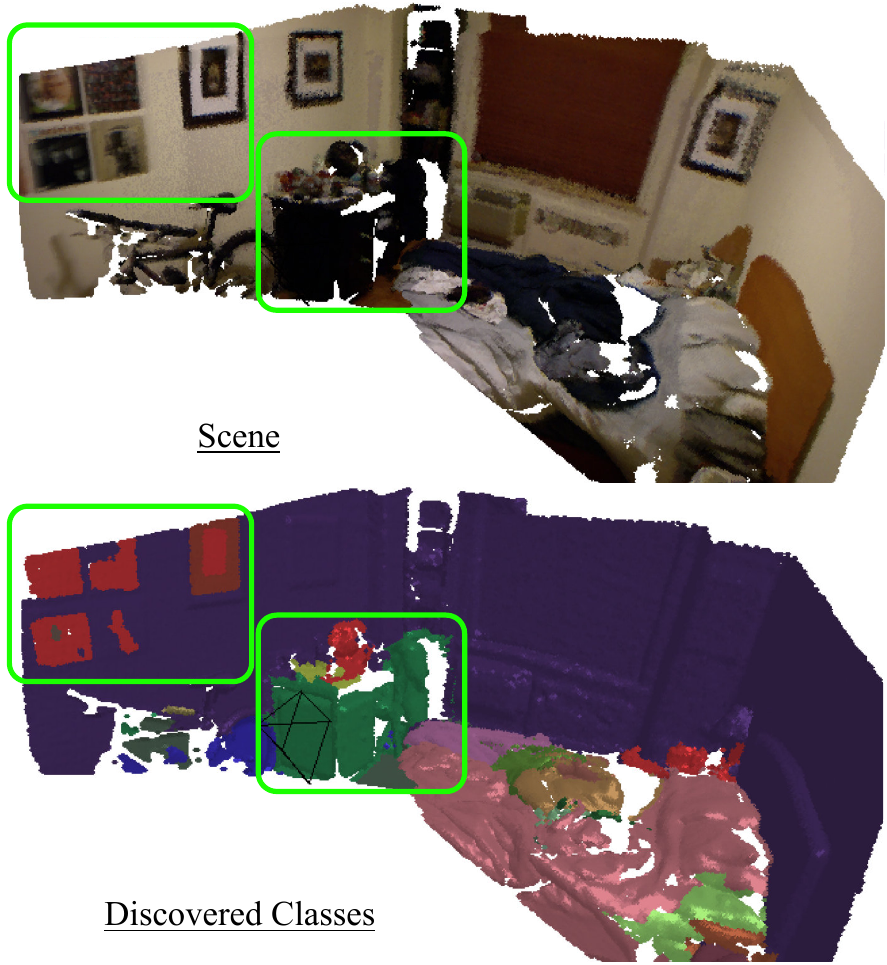}
    \caption{Proposed method incrementally discovers new classes (\eg, pictures) in the reconstructed 3D map. \label{fig:teaser}}
%    \caption{Our method incrementally discovers even classes which are excluded from the training dataset (\eg, pictures in the top row). It relies on our specifically designed geometric segmentation approach which performs more meaningful object proposals than the method of Tateno \etal~\cite{Tateno2015} (bottom row). \label{fig:teaser}}
\end{center}
\vspace{-5mm}
\end{figure}

Building a semantically annotated 3D map (\ie., semantic mapping) has become a vital research topic in computer vision and robotics communities since it provides 3D location information as well as object/scene category information. It is naturally very useful in a wide range of applications including robot navigation, mixed/virtual reality, and remote robot control.
In most of these applications, it is important to achieve both high accuracy and efficiency. Considering robot navigation, robots need to recognize objects accurately and efficiently to navigate actively changing environments without any accident. In mixed reality systems, accuracy and efficiency are important to achieve more natural interactions without delay. When controlling surgical robots remotely, they are even more essential.

Consequently, many researches have been conducted to develop an accurate and efficient system for semantic mapping~\cite{Koppula2011, Hermans2014, McCormac2017, Nakajima2018, Sengupta2013, Vineet2015, Yang2017, Kundu2014, Li2017}. Most of the recent semantic mapping systems consist of two principal components, building a 3D map from RGBD images and processing semantic segmentation on either images or the built 3D maps~\cite{Koppula2011, Hermans2014, McCormac2017, Nakajima2018}. Since the introduction of RGBD sensors such as Microsoft Kinect~\cite{Zhang2012}, many approaches have been presented for building a 3D map from RGBD images~\cite{Newcombe2011, Izadi2011, Keller2013, Lee2016}. Regarding semantic segmentation, as semantic segmentation algorithms for images have been studied in many literatures, most semantic mapping systems have adapted these algorithms. Recently, since convolutional neural networks (CNNs) improve the performance of semantic segmentation further~\cite{Long2015, Shelhamer2017, Chen2018}, CNNs have been incorporated to enhance the accuracy of semantic mapping~\cite{McCormac2017, Nakajima2018}.
%Also, additional efforts have been made to enhance the efficiency of semantic mapping systems in order to incrementally build semantic maps at every frame.
%Recently, as convolutional neural networks (CNNs) achieve state-of-the-art performances in 2D semantic segmentation tasks, many researchers incorporated the CNNs to the state-of-the-art SLAM systems to improve the accuracy of semantic mapping. Also, additional efforts have been made to enhance the efficiency of semantic mapping systems in order to incrementally build semantic maps at every frame.
%Especially, as the recent developments of convolutional neural networks (CNNs) for 2D semantic segmentation tasks, many methods efficiently combine the CNNs with the state-of-the-art SLAM method for incrementally building up a semantically annotated 3D map.

While these advancements improved the accuracy and efficiency of the overall system, the methods have limitations in the objects that the system can recognize. As previous semantic mapping systems recognize objects and scenes by training a pixel-level classifier (\eg, random forest or CNNs) using a training dataset, the systems are only able to recognize categories in the training dataset. This is a huge limitation for autonomous systems considering real-world consists of numerous objects/stuffs. Hence, we propose a novel system that can properly cluster both known objects and unseen things to enable discovering new categories. The proposed method first generates object-level segments in 3D. It then performs clustering of the object-level segments to associate objects of the same class and to discover new object classes.

The contributions of this paper are as follow: (1) We present, to the best of our knowledge, the first semantic mapping system that can properly discover clusters of both known objects and unseen objects in a 3D map (see \fref{fig:teaser}); (2) To effectively handle deep features and geometric cues in clustering, we propose to estimate the reliability of the deep features from CNNs using the entropy of the probability distribution from CNNs. We then use the estimated confidence for weighting the two types of features; (3) We propose to utilize segments instead of elements (\ie., surfel and voxel) in assigning/updating features and in clustering to efficiently reduce computational cost and space complexity. It enables the overall framework to run in semi-real-time; (4) We improve object proposals in a 3D map by utilizing both geometric and color information. It is especially important for the regions with poor geometric characteristics (\eg, pictures on a wall) ; (5) We demonstrate the effectiveness and efficiency of the proposed system by training CNNs on a subset of classes in a dataset and by discovering the other subset of classes by using the proposed method.

\section{Related Work}

\vspace{1mm}
\noindent\textbf{Semantic Scene Reconstruction}
Koppula \etal presented one of the earliest works on semantic scene reconstruction using RGBD images~\cite{Koppula2011}. Given multiple RGBD images, they first stitched the images to a single 3D point cloud. They then over-segmented the point cloud and labeled each segment using a graphical model.

As many 2D semantic segmentation approaches achieved impressive results~\cite{Long2015, Shelhamer2017, Chen2018}, Hermans \etal proposed to use 2D semantic segmentation for 3D semantic reconstruction instead of segmenting 3D point clouds~\cite{Hermans2014}. They first processed 2D semantic segmentation using randomized decision forests (RDF) and refined the result using a dense Conditional Random Fields (CRF). They then transferred class labels to 3D maps. Since, recently, convolutional neural networks (CNNs) further improved 2D semantic segmentation, McCormac \etal presented a system that utilizes CNNs for 2D semantic segmentation instead of RDF~\cite{McCormac2017}. While we focus on semantic scene reconstruction methods using RGBD images, there are methods using stereo image pairs~\cite{Sengupta2013, Vineet2015, Yang2017} and using a monocular camera~\cite{Kundu2014, Li2017}.

While all the previous works~\cite{Koppula2011, Hermans2014, McCormac2017, Sengupta2013, Vineet2015, Yang2017, Kundu2014, Li2017} can recognize only learned object classes, we propose, to the best of our knowledge, the first semantic scene reconstruction system that can segment unseen object classes as well as trained classes. 

\vspace{1mm}
\noindent\textbf{Image Segmentation}
Image segmentation has been studied in many literatures~\cite{Ray1999, Shi2000, Boykov2001, Deng2001, Comaniciu2002, Felzenszwalb2004, Huang2004, Fulkerson2012, Arbelaez2009, Arbelaez2011}. Relatively recently, Pont-Tuset \etal proposed an approach for bottom-up hierarchical image segmentation~\cite{Pont2017}. They developed a fast normalized cuts algorithm and also proposed a hierarchical segmenter that uses multiscale information. They then employed a grouping strategy that combines multiscale regions into highly-accurate object proposals. As convolutional neural networks (CNNs) have become a popular approach in semantic segmentation, Xia \etal proposed a CNN-based method for unsupervised image segmentation~\cite{Xia2017}. They segmented images by learning autoencoders with the consideration of the normalized cut and smoothed the segmentation outputs using a conditional random field. They then processed hierarchical segmentation that first converts over-segmented partitions into weighted boundary maps and then merges the most similar regions iteratively. 
 
Considering RGBD data, Yang \etal proposed a two-stage segmentation method that consists of over-segmentation using 3-D geometry enhanced superpixels and graph-based merging~\cite{Yang2015}. They first applied a K-means-like clustering method to the RGBD data for over-segmentation using an 8-D distance metric constructed from both color and 3-D geometrical information. They then employed a graph-based model to relabel the superpixels into segments considering RGBD proximity, texture similarity, boundary continuity, and the number of labels.

Comparing to the previous works~\cite{Ray1999, Shi2000, Boykov2001, Deng2001, Comaniciu2002, Felzenszwalb2004, Huang2004, Fulkerson2012, Arbelaez2009, Arbelaez2011, Pont2017, Xia2017, Yang2015}, this work differs from them in two aspects. First, we propose a segmentation algorithm for 3D reconstructed scenes rather than images. Second, we aim to group pixels with the same semantic meaning to a cluster even if they are distant or separated by another segment.  

%\begin{figure*}[t]
%\begin{center}
%    \includegraphics[width=1.0\hsize]{flow.pdf}
%    \caption{Flow of the proposed framework. \label{fig:flow}}
%\end{center}
%\end{figure*}

\begin{figure}[t]
\begin{center}
    \includegraphics[width=1.0\hsize]{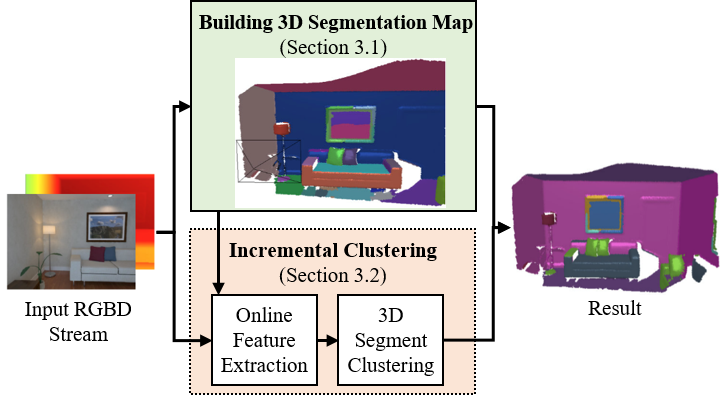}
    \caption{Overview of the proposed framework. \label{fig:flow}}
\end{center}
\vspace{-5mm}
\end{figure}

\section{Class Discovery for Semantic Segmentation}\label{sec:method}

% Motivation
In order to discover new classes of semantic segments, we need a method for aggregating and clustering unknown segments (\ie, segments of the image which cannot be classified into a known class). A central component of our proposed approach is the segmentation of a dense 3D reconstruction of the scene, which we call the \textit{3D segmentation map}, which is used to aggregate information about each 2D image segment and that information is used to perform the 3D segment clustering to discover new `semantic' (a nameless category) classes.

% Overview
To incrementally discover object classes using RGBD sensing, we first propose to build a 3D segmentation map for object-level segmentation in 3D. Second, we perform clustering of the object-level segments to associate objects of the same class and to discover new object classes. \fref{fig:flow} shows the overview of the proposed framework. Given an input RGBD stream, we build a 3D segmentation map (\sref{sec:inseg}) and process incremental clustering (\sref{sec:inc_clustering}). The incremental clustering consists of extracting features for each frame (\sref{sec:fext}) and clustering using the features (\sref{sec:omcl}). The output of the proposed method is the visualization of clustering membership on a reconstructed 3D map.

\subsection{Building the 3D Segmentation Map} \label{sec:inseg}

As mentioned above, the 3D segmentation map is the key data structure which is used to aggregate information about 2D image segmentation to discover new semantic classes. Building the 3D segmentation map is an incremental process, which consists of the following four processes applied to each frame: (1) SLAM for dense 3D map reconstruction; (2) SLIC for superpixel segmentation; (3) Agglomerative clustering; and (4) Updating the 3D segmentation map. We describe the details of each processing step below.

%Building a 3D segmentation map is a key component of the proposed method as it is used to discover object clusters by identifying coherent regions in the map. An image of the 3D segmentation map is shown in Figure \ref{}. We will show how we utilize the 3D segmentation map to reduce the computational complexity and memory usage of the proposed method. Since processing at the element-level (\eg, surfels and voxels) is expensive, we propose a novel approach that assigns/updates features at for each 3D segment and clusters them to discover new classes. As the 3D segmentation map is used to obtain object boundaries and to cluster objects, building the map with high accuracy is crucial. Hence, we improve the quality of segmentation by properly utilizing both color information and depth cues. This contributes to overcoming the limitations of the previous methods~\cite{Tateno2015, finman2013toward} which only uses depth information.

%As shown in~\fref{fig:segmentation}, given an input RGBD stream, we incrementally build a 3D segmentation map where each surfel has an assigned segmentation label.

\begin{figure}[t]
\begin{center}
    \includegraphics[width=1.0\hsize]{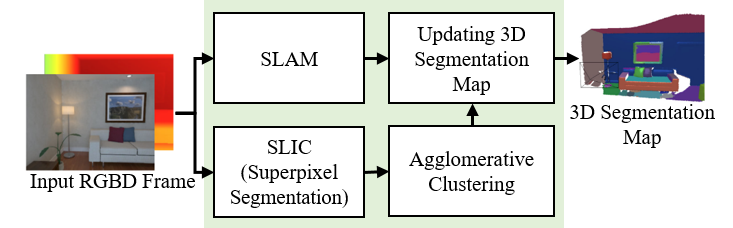}
    \caption{\textbf{Building 3D Segmentation Map.}  The output of this processing is object-level segments in 3D. We build the 3D map by propagating 2D segmentation to the existing 3D segmentation map. (\sref{sec:inseg}). \label{fig:segmentation}}
\end{center}
\vspace{-5mm}
\end{figure}

%The 3D segmentation map is a key component of the proposed method as it is used to discover object clusters by identifying coherent regions in the map.

%Moreover, we utilize the 3D segmentation map to reduce the computational complexity and memory use of the proposed method. As processing feature updates or clustering at element-level (\eg, surfels and voxels) is expensive, we propose a novel approach that assigns/updates features at each segment and clusters the segments.

%Furthermore, comparing to previous methods~\cite{Tateno2015, finman2013toward}, we improve the quality of segmentation by utilizing both color information and depth cues. As previous methods only depend on depth information, they have limitations in segmenting regions where geometric features are poor.

%In the following paragraphs, we describe the four processing stages: (1) SLAM, (2) SLIC, (3) Superpixel clustering, and (4) updating 3D segmentation map.

\vspace{1mm}
\noindent\textbf{Dense SLAM.} In order to estimate camera poses and incrementally build a 3D map, we employ the dense SLAM method, InfiniTAM v3~\cite{Prisacariu2017}.
The method builds 3D maps using an efficient and scalable representation method which was proposed by Keller \etal~\cite{Keller2013}. The representation is a point-based description with normal information and is referred to as {\it surfel}. We denote surfels using $s_k$.

The \textit{surfel} is a fundamental element in our reconstructed 3D map (like pixels on an image). Given a new depth frame, we generate surfels and fuse them into the existing reconstructed 3D map. Hence, building the 3D segmentation map includes building a reconstructed 3D map using SLAM and grouping surfels in the reconstructed 3D map.

%At the $t$-th incoming RGBD frame, the method estimates the current camera pose $\bm{T}_t \in \mathbb{SE} (3)$ using an Iterative Closest Point algorithm~\cite{Low2004} and RGB alignment. Then, the new surfels are generated from the current depth map and are fused into the existing surfels of the 3D map using the estimated camera pose. While fusing, it refines 3D coordinates and normals considering both the existing surfels and the newly generated surfels.

\vspace{1mm}
\noindent\textbf{RGBD SLIC.} For every RGBD frame, we first implement a modified SLIC superpixel segmentation algorithm to generate roughly 250 superpixels (small image regions) for each frame. To use both color information and geometric cues, we define a new distance metric $D_s$ that uses the color image $\mathcal{I}^{lab}_t(\bm{u}) \in \mathbb{R}^3$ in the CIELAB color space, the normal map $\mathcal{N}_t(\bm{u}) \in \mathbb{R}^3$, and the image coordinates $\bm{u} = (x, y) \in \mathbb{Z}^2$.
The distance $D_s$ between pixels $\bm{u}$ and $\bm{v}$ is computed as follows:
\begin{equation}
\label{eq:sp_dist}
\begin{split}
& D_s = d_{lab} + \alpha d_{n} + \beta d_{xy} , \\
& d_{lab} = || \mathcal{I}^{lab}_t(\bm{u}) - \mathcal{I}^{lab}_t(\bm{v}) ||_2 , \\
& d_{n} =  || \mathcal{N}_t(\bm{u}) - \mathcal{N}_t(\bm{v}) ||_2 , \\
& d_{xy} = || \bm{u} - \bm{v} ||_2 ,
\end{split}
\end{equation}
where $\alpha$ and $\beta$ are constants for weighting $d_{n}$ and $d_{xy}$. Given the set of superpixels from the SLIC segmentation, we compute the averaged color $\bm{c}^{lab} \in \mathbb{R}^3$, vertex $\bm{v} \in \mathbb{R}^3$, and normal $\bm{n} \in \mathbb{R}^3$ of each superpixel $r$, which will be used to further merge superpixels into bigger 2D regions.

\vspace{1mm}
\noindent\textbf{Agglomerative Clustering.} Since the SLIC superpixel segmentation tends to generate a grid of segments with similar sizes, we perform agglomerative clustering and merging to produce object-level segments. The clustering and merging are based on the similarity in $\bm{c}^{lab}$, $\bm{v}$, and $\bm{n}$ between superpixels. Specifically, we compute the similarity $\Lambda$ in color space, the geometric distance $\Psi$ in 3D space, and convexity $\Phi$ in shape. We then merge the superpixels if all the measured similarity/distances meet the following conditions.

Consider two neighboring superpixels $(r_a, r_b)$. The $\Lambda$, $\Psi$, and $\Phi$ are computed as follow:
\begin{equation}
\begin{split}
\label{eq:phi}
& \Lambda(r_a,r_b) = || \bm{c}_a - \bm{c}_b ||_2 , \\
& \Psi(r_a,r_b) = || (\bm{v}_b - \bm{v}_a) \cdot \bm{n}_a ||_2 , \\
& \Phi(r_a,r_b) =
    \left\{%
        \begin{array}{ll}
1 & \text{if } (\bm{v}_b - \bm{v}_a) \cdot \bm{n}_a > 0 , \\
\bm{n}_a \cdot \bm{n}_b & \mbox{otherwise.}
        \end{array}%
    \right.
\end{split}
\end{equation}
%Specifically, $\Phi$ outputs value 1 if the superpixels are on a convex shape and the dot product $\bm{n}_a \cdot \bm{n}_b$ if the superpixels are on a concave shape. 1 is the maximum value of a dot product between normals.
Given $\Lambda$, $\Psi$, and $\Phi$, the pair of superpixels $(r_a, r_b)$ are merged only when they satisfy the predetermined criteria:
\begin{equation}
\Lambda < \sigma_{\Lambda} \text{ and } \Psi < \sigma_{\Psi} \text{ and } \Phi > \sigma_{\Phi} ,
\end{equation}
where $\sigma_{\Lambda}$, $\sigma_{\Psi}$, and $\sigma_{\Phi}$ denote the corresponding thresholds for $\Lambda$, $\Psi$, and $\Phi$, respectively. Regarding convexity criteria, it is based on the observation that objects on captured images usually have convex shapes \cite{Tateno2015}. Consequently, we penalize merging regions with concave shapes.
$\sigma_{\Psi}$ is computed using the noise model in \cite{nguyen2012modeling}, which presented the relationship between noise and distance from a sensor.

\vspace{1mm}
\noindent\textbf{3D Segmentation Map Update.} Given the 2D segmentation result of current frame, we update the 3D segmentation map. We employ the efficient and scalable segment propagation method in~\cite{Tateno2015} to assign/update a segment label $l_i$ to each surfel $s_k$.

%================================================================%
\subsection{Incremental Clustering} \label{sec:inc_clustering}

\begin{figure}[t]
\begin{center}
    \includegraphics[width=1.0\hsize]{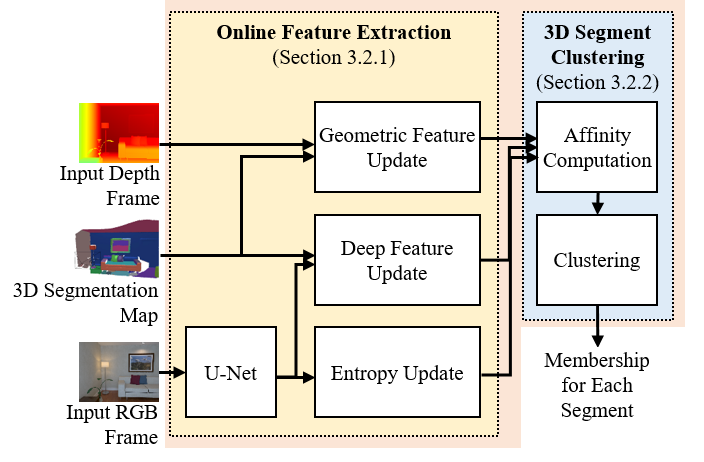}
    \caption{\textbf{Incremental 3D Segment Clustering.} This clustering is to associates objects of the same class or to discover new classes using object-level segments in the 3D segmentation map. (\sref{sec:inc_clustering}). \label{fig:clustering}}
\end{center}
\vspace{-5mm}
\end{figure}

In the previous section, we generate object-level segments by clustering and merging superpixels. The object-level segments are then used to update the 3D segmentation map. Given the object-level segments in the 3D segmentation map, incremental clustering aims to discover new object classes by clustering the object-level segments. To cluster the segments, we first extract features using an input RGBD frame and the 3D segmentation map. We then cluster by computing weighted similarity between the segments. We describe the details of online feature extraction in~\sref{sec:fext} and those of 3D segment clustering in~\sref{sec:omcl} (also, see~\fref{fig:clustering}).

\subsubsection{Online Feature Extraction} \label{sec:fext}
%\textcolor{blue}{(KK: Need motivation paragraph here. Not a description of the algorithm!)}

In order to accurately associate objects of the same class or to discover new object classes, we need a method for estimating similarity between object segments in the 3D segmentation map. While measuring similarity can be as simple as computing distance in color space, more meaningful measurement is required to accurately determine object classes. Moreover, as objects often appear on multiple frames in a consecutive video, we can improve the similarity measurement by utilizing previous frames. Lastly, as record-keeping all the information from previous frames is expensive, we need an efficient method to store the past information.

%we need to measure it in a space that separates objects of different classes, but brings together those of same class. Moreover, we prefers features that are translation
%semantic meanings,

To estimate more meaningful similarity, we utilize both features from color images and geometric features as they are often complementary.
Especially, as convolutional neural networks have achieved impressive results in per-pixel classification tasks~\cite{Long2015, Shelhamer2017, Chen2018}, we extract features from color images using CNNs. The extracted deep features and geometric features for each frame are then used to update the features for each segment in the 3D segmentation map. By aggregating the features from all the previous frames, we improve the robustness of the features in the 3D segmentation map. Moreover, storing/updating the features for each segment is a very effective strategy for both saving memory usage and reducing computations for 3D segment clustering. Considering the number of segments is much smaller than that of surfels in the 3D map, the reduction in memory usages is very significant. Specifically, the memory usage is reduced from $O(N_s·(S + G + 1))$ to $O(N_l·(S + G + 1))$ where $N_s$ and $N_l$ denote the number of surfels and that of object-level segments in the 3D segmentation map, respectively; $S$ and $G$ represent the dimension of the deep features and that of the geometric features, respectively.

%In our approach, deep features $\bm{f}^{\text{CNN}}_{l_i} \in \mathbb{R}^{S}$, geometric features $\bm{f}^{\text{GEO}}_{l_i} \in \mathbb{R}^{G}$, and entropy $e_{l_i} \in \mathbb{R}$ are extracted/assigned/updated for each segment $l_i$.
%Hence, the space complexity for storing the two types of features and entropy is $O(N_l·(S + G + 1))$ where $N_l$, $S$, and $G$ denote the number of segments in the 3D segmentation map, the dimension of the deep features, and the dimension of the geometric features, respectively.
%Thus, the strategy of assigning/updating features region-wise reduces memory demands effectively.

While CNNs have shown impressive results, the reliability of deep features can vary depending on the region of the input image. We hypothesize that the regions that CNNs can predict a class with high confidence, can be clustered accurately using deep features. Hence, we estimate the reliability of deep features using predicted probability distribution from CNNs. Specifically, we compute the confidence by calculating the entropy of the predicted probability distribution. We then, based on the estimated reliability, compute weighted affinity using the similarity of geometric feature and that of deep features between object-level segments in the 3D segmentation map.

For deep features and entropy, we employ the U-Net architecture~\cite{Ronneberger2015} since our target applications (\eg robot navigation) often demand short processing time. The network takes only $36 ms$ to process an input image of $320 \times 240$ resolution.
Also, by using the same network for both processing, we can save computations.

%, we can improve the robustness of extracted features at a frame by accumulating the information from previous frames. Hence, for the regions that appeared in previous frames, we combine the extracted features from the current frame and those from the previous frames.

%However, as record-keeping all the features from all the previous frames is expensive, we only hold an accumulated features in the geometric 3D map and update the features using that from the current frame. Also, as processing features at element-level (\eg surfels and voxels) is expensive, we assign semantic and geometric features to each region in the geometric 3D map.

%In order to make clusters in object-level, not element-level (\eg, surfels and voxels), we assign semantic and geometric features to each segmentation label $l_i$ associated to each region constituting the geometric 3D map.

%We describe the process of extracting the per-frame deep features, geometric features, and entropy from the incoming RGBD frames. We first present the method of extracting deep features and estimating entropy using convolutional neural networks. We then describe extracting geometric features for each region on the current image plane.

%As shown in~\fref{fig:clustering}, given the network and input images, we extract entropy and deep feature map. The entropy is computed to estimate the confidence of deep features and is used for feature updates in the following section. The deep feature is extracted to utilize in clustering.

\vspace{1mm}
\noindent\textbf{Geometric Feature Extraction/Update.} %\label{sec:geof}
To extract translation/rotation-invariant and noise-robust geometric features, we first estimate a Local Reference Frame (LRF) for each segment. We then extract geometric features for each segment using a fast and unique geometric feature descriptor, Global Orthographic Object Descriptor (GOOD)~\cite{kasaei2016good}.

%To extract geometric features for each region on the current image frame, we first render the updated 3D segmentation map onto the current image plane using the estimated camera pose $\bm{T}_t$ and the 3D position $\bm{x}(k)$ associated with each surfel $s_k$.
%The rendered segmentation map $\mathcal{R}(\bm{u})$, where each component is associated to a segmentation label $l_i$, is generated with $\mathcal{R}(\pi(\bm{T}^{-1}_t \bm{x}(k))) = l_i(k)$ by denoting the segmentation label $l_i$ of a surfel $s_k$ with $l_i (k)$.
%We then extract geometric features $\mathcal{F}^{\text{GEO}}_t(l_i) \in \mathbb{R}^{G}$ for each segment $l_i$ by first making translation/rotation-invariant and noise-robust and by utilizing a geometric feature descriptor.

%To make geometric features invariant to translations/rotations and robust to noise, we estimate a Local Reference Frame (LRF) $(X, Y, Z)$ for each segment $l_i$.
%To estimate LRF, we first compute the three principal axes $(X, Y, Z)$ for each segment $l_i$ using the Principal Component Analysis (PCA).
Given a depth map, to estimate LRF for each segment, we need the 3D segmentation map on the current image plane. Hence, we first render the segmentation map to the current image plane and obtain the rendered segmentation map $\mathcal{R}$ with segment labels $l_i$.
We then compute the LRF by processing the Principal Component Analysis (PCA) for each segment.
In more details about processing the PCA, we first compute the normalized covariance matrix and then perform eigenvalue decomposition.
The normalized covariance matrix $\bm{C}_{l_i}$ of each segment $l_i$ is computed using the vertex map $\mathcal{V}_t$ and the rendered segmentation map $\mathcal{R}$ as follows:
\begin{equation}
\begin{split}
\bm{C}_{l_i} &= \frac{1}{|\mathcal{U}_{l_i}|} \sum_{\bm{v} \in \mathcal{U}_{l_i}} (\bm{v} - \bm{o}_{l_i})(\bm{v} - \bm{o}_{l_i})^T , \\
\bm{o}_{l_i} &= \frac{1}{|\mathcal{U}_{l_i}|} \sum_{\bm{v} \in \mathcal{U}_{l_i}} \bm{v} , \\
\mathcal{U}_{l_i} &= \{\mathcal{V}_t(\bm{u}) | \mathcal{R}(\bm{u}) = l_i \} ,
\end{split}
\end{equation}
where $\bm{o}_{l_i}$ represents the geometric center of the segment $l_i$; $\mathcal{U}_{l_i}$ denotes the set of vertices that belong to the segment $l_i$ on the current frame; $| \cdot |$ represents the number of elements in the set.
%The geometric center $\bm{o}_{l_i}$ of each segment $l_i$ is computed using the vertex map $\mathcal{V}_t$ and the rendered segmentation map $\mathcal{R}$ as follows:
%\begin{equation}
%\bm{o}_{l_i} = \frac{1}{|\mathcal{U}_{l_i}|} \sum_{\bm{v} \in \mathcal{U}_{l_i}} \bm{v} ,
%\end{equation}
%\begin{equation}
%\mathcal{U}_{l_i} = \{\mathcal{V}_t(\bm{u}) | \mathcal{R}(\bm{u}) = l_i \} ,
%\end{equation}
%where $\mathcal{U}_{l_i}$ denotes the set of vertices that belong to the segment $l_i$ on the current frame; $| \cdot |$ represents the number of elements in the set.
%Using $\bm{o}_{l_i}$, the normalized covariance matrix $\bm{C}_{l_i}$ is computed for each $l_i$ as follows:
%\begin{equation}
%\bm{C}_{l_i} = \frac{1}{|\mathcal{U}_{l_i}|} \sum_{\bm{v} \in \mathcal{U}_{l_i}} (\bm{v} - %\bm{o}_{l_i})(\bm{v} - \bm{o}_{l_i})^T .
%\end{equation}
We then perform eigenvalue decomposition on $\bm{C}_{l_i}$ as follows:
\begin{equation}
\bm{C}_{l_i}\bm{X}_{l_i} = \bm{E}_{l_i}\bm{X}_{l_i},
\end{equation}
where $\bm{X}_{l_i}$ is a matrix with three eigenvectors; $\bm{E}_{l_i} = diag(\lambda_1, \lambda_2, \lambda_3)$ is a diagonal matrix with the corresponding eigenvalues.
$\bm{X}_{l_i}$ is directly utilized as the LRF.
%$(X, Y, Z)$.
%where $\bm{X}_{l_i}$ consists of three eigenvectors and $\bm{E}_{l_i} = diag(\lambda_1, \lambda_2, \lambda_3)$ is a diagonal matrix with the corresponding eigenvalues.
%We directly employ $\bm{X}_{l_i} = [\bm{x}_1, \bm{x}_2, \bm{x}_3]$ as the LRF $(X, Y, Z)$.

Lastly, we employ a fast and unique geometric feature descriptor, GOOD~\cite{kasaei2016good}.
For each $l_i$, we transform the set of vertices $\mathcal{U}_{l_i}$ using the LRF. We then fed the transformed vertices into the descriptor to obtain the frame-wise geometric feature $\mathcal{F}^{\text{GEO}}_t(l_i) \in \mathbb{R}^{75}$.

After computing $\mathcal{F}^{\text{GEO}}_t(l_i)$ using the current depth map, the geometric features $\bm{f}^{\text{GEO}}_{l_i}$ in the 3D segmentation map are updated as follows:
\begin{equation}
\begin{split}
&    \bm{f}^{\text{GEO}}_{l_i} \gets \frac{1}{Z^{\text{GEO}}_{l_i}} \cdot \frac{\Omega \bm{f}^{\text{GEO}}_{l_i} + \mathcal{F}^{\text{GEO}}_t(l_i)}{\Omega + 1}, \\
&    \Omega \gets \Omega + 1 .
\end{split}
\end{equation}
This updates are applied to all segments $l_i$ on the rendered segmentation map $\mathcal{R}$.
$Z^{\text{GEO}}_{l_i}$ denotes the constant for normalizing the feature vector $\bm{f}^{\text{GEO}}_{l_i}$.

\vspace{1mm}
\noindent\textbf{Deep Feature Extraction/Update.}
We utilize the output of the layer just before the last classification layer for deep feature map.
The per-frame deep feature map is denoted as $\mathcal{F}^{\text{CNN}}_t(\bm{u}) \in \mathbb{R}^{S}$. The size of $\mathcal{F}^{\text{CNN}}_t$ is $W \times H \times S$ where $W$ and $H$ represent the width and height of an input image, respectively; and $S$ denotes the number of channels (\ie. the dimension of the features) which is $64$.
%The U-Net, which is the CNN architecture we employ in this work, takes $S$ as $64$.
%At the same time, we utilize the outputs of the layer just before the last layer as the deep feature map.
%In the network architecture of the U-Net, the layer outputs the feature map in the size of $W \times H \times 64$, by letting W and H as the size of input image respectively.
%Thus, we can denote the frame-wise deep feature map as FCNN(u)t = fCNN < R64.

We update the deep features $\bm{f}^{\text{CNN}}_{l_i}$ for each segment $l_i$ in the 3D segmentation map by employing incremental averaging approach and by using the per-frame deep features.
Since deep features and entropy are extracted for each pixel while geometric features are obtained for each segment $l_i$, the procedures for updating are slightly different.
The deep features $\bm{f}^{\text{CNN}}_{l_i}$ are updated as follows:
\begin{equation}
\label{eq:cnn}
\begin{split}
&    \bm{f}^{\text{CNN}}_{l_i = \mathcal{R}(\bm{u})} \gets \frac{1}{Z^{\text{CNN}}_{l_i}} \cdot \frac{\Gamma \bm{f}^{\text{CNN}}_{l_i = \mathcal{R}(\bm{u})} + \mathcal{F}^{\text{CNN}}_t(\bm{u})}{\Gamma + 1} ,\\
&    \Gamma \gets \Gamma + 1,
\end{split}
\end{equation}
where $Z^{\text{CNN}}_{l_i}$ is the normalizing constant for $\bm{f}^{\text{CNN}}_{l_i}$; $\bm{u}$ is all the coordinates on $\mathcal{F}^{\text{CNN}}_t$.

\vspace{1mm}
\noindent\textbf{Entropy Computation/Update.}
The entropy is computed by first estimating the probability distribution for each class and by measuring the Shannon entropy~\cite{Shannon1948} using the probability distribution. As the network is trained for semantic segmentation, the probability distribution is obtained by the output of the softmax layer of the network. The entropy $\mathcal{E}(\bm{u}) \in \mathbb{R}$ is computed at each pixel $\bm{u}$ as follows:
\begin{equation}
\mathcal{E}(\bm{u}) = - \sum_c P_c(\bm{u}) \log P_c(\bm{u}) ,
\end{equation}
where $P_c(\bm{u}) \in \mathbb{R}$ is the probability for the class $c$ at the pixel $\bm{u}$.
%We feed the input frame It to the U-Net to obtain the probability distributions P(It(u) = c|It) for each pixel u.
%Next, we obtain the entropy map E(u) = e < R for each pixel through the probability distributions P(It(u) = c|It) by following the definition of the Shannon entropy \cite{}.
Then, $\mathcal{E}(\bm{u})$ is used for updating the entropy $e_{l_i}$ for each segment $l_i$ in the 3D segmentation map as follows:
\begin{equation}
\label{eq:en}
\begin{split}
&    e_{l_i = \mathcal{R}(\bm{u})} \gets \frac{\Gamma e_{l_i = \mathcal{R}(\bm{u})} + \mathcal{E}(\bm{u})}{\Gamma + 1} , \\
&    \Gamma \gets \Gamma + 1,
\end{split}
\end{equation}
where $\bm{u}$ is all the coordinates on $\mathcal{E}$.

\begin{table*}[t]
    \begin{center}
   \caption{Quantitative comparison on the NYUDv2 dataset~\cite{Silberman2012}. Supervised methods versus unsupervised methods (ours). \label{tab:iou}}
   %\caption{Quantitative results for the NYUv2 dataset \cite{Silberman2012} with IoU measurement. The classes with red text are eliminated from the dataset used for training U-Net. All accuracy evaluations were performed at 320 $\times$ 240 resolution.  \textcolor{blue}{(KK: Add citations inside table)} \label{tab:iou}}
      \scalebox{0.81}{
       \begin{tabular}{l|ccccccccc|cccc|c} \bhline{1pt}
           & 
          \multicolumn{1}{|c|}{ \cellcolor{bed}} & 
          \multicolumn{1}{c|}{  \cellcolor{books}} & 
          \multicolumn{1}{c|}{  \cellcolor{chair}} & 
          \multicolumn{1}{c|}{  \cellcolor{floor}} & 
          \multicolumn{1}{c|}{  \cellcolor{furniture}} & 
          \multicolumn{1}{c|}{  \cellcolor{objects}} & 
          \multicolumn{1}{c|}{  \cellcolor{sofa}} & 
          \multicolumn{1}{c|}{  \cellcolor{table}} & 
          \multicolumn{1}{c|}{  \cellcolor{wall}} & 
          \multicolumn{1}{c|}{  \cellcolor{ceiling}} & 
          \multicolumn{1}{c|}{  \cellcolor{painting}} & 
          \multicolumn{1}{c|}{  \cellcolor{tv}} & 
          \multicolumn{1}{c|}{  \cellcolor{window}} &  \\
          Method & \multicolumn{9}{c|}{classes in training dataset} & \multicolumn{4}{c|}{novel classes} & \bf mean \\
          \cline{2-14}
           & 
          \multicolumn{1}{c}{bed} & 
          \multicolumn{1}{c}{book} & 
          \multicolumn{1}{c}{chair} & 
          \multicolumn{1}{c}{floor} & 
          \multicolumn{1}{c}{furn.} & 
          \multicolumn{1}{c}{obj.} & 
          \multicolumn{1}{c}{sofa} & 
          \multicolumn{1}{c}{table} & 
          \multicolumn{1}{c|}{wall} & 
          \multicolumn{1}{c}{ceil.} & 
          \multicolumn{1}{c}{pict.} & 
          \multicolumn{1}{c}{tv} & 
          \multicolumn{1}{c|}{wind.} & 
          {\bf IoU\ } \\ \hline
          U-Net~\cite{Ronneberger2015}       & 50.32 & 22.42 & 36.55 & 55.62 & 36.85 & 27.27 & 48.44 & 33.78 & 55.14 &     - &     - &     - &     - & - \\
          Nakajima \etal~\cite{Nakajima2018} & 62.82 & \textbf{27.27} & \textbf{42.56} & {\bf 68.43} & 44.62 & 24.63 & 45.04 & \textbf{42.30} & 26.82 &     - &     - &     - &     - & - \\
          \hline
          {\bf Ours + 3D Map ~\cite{Tateno2015}} & 62.80 & 23.96 & 33.10 & 63.41 & 50.58 & 27.28 & {\bf 58.68} & 40.23 & 54.53 & {\bf 31.42} & 19.37 & 43.98 & 31.30 & 41.59  \\
          {\bf Ours}                                & {\bf 64.22} & 22.28 & 41.79 & 67.38 & {\bf 56.15} & \textbf{28.61} & 49.31 & 40.95 & {\bf 63.18} & 29.30 & {\bf 28.69} & {\bf 52.20} & {\bf 53.92} & {\bf 46.05} \\ 
          \bhline{1pt}
       \end{tabular}
    }
    \end{center}
 \end{table*}

\subsubsection{3D Segment Clustering} \label{sec:omcl}
Given semantic and geometric features in the 3D segmentation map from the feature updating stage, we apply a graph-based unsupervised clustering algorithm to cluster regions in the 3D segmentation map. We specifically employ the Markov clustering algorithm (MCL)~\cite{VanDongen2000} because of the flexible number of clusters and computational cost. Since we aim to be able to handle unknown objects in a scene, we need the number of clusters (class categories) to be flexible, like the MCL. Furthermore, since the computational cost $O(M^3)$ of the MCL comes from the multiplication of two matrices with the size $M \times M$, where $M$ denotes the number of nodes in the graph, the cost can be turned into $O(M)$ by parallelizing the processing in a GPU. Accordingly, it reduces processing time and makes more appropriate for an online system.

%The goal of this stage is to update clusters for each region in the geometric 3D map, with both of semantic and geometric features assigned and updated through the Feature Update stage.
%In this work, we employ the Markov clustering algorithm (MCL)~\cite{VanDongen2000}, an unsupervised cluster algorithm for graphs.
%Considering the aim of this work: handling unknown objects in the scene where the number of class categories is not fixed, such a hierarchical clustering method is preferable.
%Furthermore, the computational cost for MCL is $O(M^3)$ by denoting $M$ the number of nodes.
%Since the cost $O(M^3)$ comes from the multiplication for the matrix with the size $M \times M$, the cost can be turned into $O(M)$ with the GPU implementation, and thus being durable on the online system.

We define the similarity $s(i, j)$ between nodes (\ie. regions $l_i$ and $l_j$ in the 3D segmentation map).
The weight values $w_i$ and $w_j$ are first computed using the entropy $e$ and the number $N$ of classes in the training dataset for the U-Net as follows:
\begin{equation}
    w_i = \frac{e_{l_i}}{\log N},\  w_j = \frac{e_{l_j}}{\log N} .
\end{equation}
The denominator $\log N$ is selected to make $w$ to be in [0,1] considering the maximum value of $e_{l_i}$ is $\log N$.
%where $N$ denotes the number of classes in the training dataset for the U-Net, and \textcolor{red}{thus $\log N$ is the maximum value of $e_{l_i}$}.
The similarity $s(i, j)$ is then defined using $w_i$ and $w_j$ as follows:
%We define the similarity $s(i, j)$ between nodes, \ie., regions $l_i$ and $l_j$ in the geometric 3D map.
%First, we calculate the weight values $w_i$ and $w_j$ as follows:
%\begin{equation}
%    w_i = \frac{e_{l_i}}{\log N},\  w_j = \frac{e_{l_j}}{\log N}.
%\end{equation}
%$N$ denotes the number of classes in the dataset used for training U-Net, and thus $\log N$ is the maximum value of $e_{l_i}$.
%With weight values $w_i$ and $w_j$, the similarity $s(i, j)$ is defined with the constant $\eta$ as follows:
\begin{equation}
\label{eq:sim}
\begin{split}
s(i, j) &= e^{-\eta d(i,j)}, \\
d(i, j) &= || (1-w_i)\bm{f}^{\text{CNN}}_{l_i} - (1-w_j)\bm{f}^{\text{CNN}}_{l_j} ||_2 \\
& + || w_i\bm{f}^{\text{GEO}}_{l_i} - w_j\bm{f}^{\text{GEO}}_{l_j} ||_2 ,
\end{split}
\end{equation}
where $\eta$ is a predefined constant.
Based on the assumption that the entropies of regions belonging to unknown object categories are high, the similarity measurement between these regions is more relying on geometric features than deep features.
We calculate the similarity $s(i, j)$ for each pair of region $(i, j)$ and feed the similarities to the MCL to update clusters.

\section{Experiments and Results}

\begin{table*}[t]
\begin{center}
   \caption{Ablation study on effects of deep features and geometric features for clustering.   \label{tab:ablation}}
   \scalebox{0.81}{
       \begin{tabu} to 1.22\textwidth {X[l,m]|ccccccccc|cccc|c} 
       \bhline{1pt}
          \multirow{2}{*}{Method} & \multicolumn{9}{c|}{classes in training dataset} & \multicolumn{4}{c|}{novel classes} & \\
          \cline{2-14}
           & 
          \multicolumn{1}{c}{bed} & 
          \multicolumn{1}{c}{book} & 
          \multicolumn{1}{c}{chair} & 
          \multicolumn{1}{c}{floor} & 
          \multicolumn{1}{c}{furn.} & 
          \multicolumn{1}{c}{obj.} & 
          \multicolumn{1}{c}{sofa} & 
          \multicolumn{1}{c}{table} & 
          \multicolumn{1}{c|}{wall} & 
          \multicolumn{1}{c}{ceil.} & 
          \multicolumn{1}{c}{pict.} & 
          \multicolumn{1}{c}{tv} & 
          \multicolumn{1}{c|}{wind.} & 
          {\bf mean IoU\ } \\ \hline
          Ours GEO-only                             & 51.95 & 21.47 & 35.99 & 64.75 & 50.28 & 28.36 & 48.98 & 39.14 & 55.80 & \textbf{29.76} & 25.38 & 44.88 & 52.43 & 42.24 \\
          Ours CNN-only                             & 60.07 & {\bf 28.23} & 37.55 & 63.53 & 49.48 & {\bf 30.16} & \textbf{51.21} & {\bf 43.59} & 59.94 & 20.82 & 22.60 & 39.41 & 42.30 & 42.22 \\
          {\bf Ours}                                & {\bf 64.22} & 22.28 & {\bf 41.79} & \textbf{67.38} & {\bf 56.15} & 28.61 & 49.31 & 40.95 & {\bf 63.18} & 29.30 & {\bf 28.69} & {\bf 52.20} & {\bf 53.92} & {\bf 46.05} \\ 
          \bhline{1pt}
       \end{tabu}
    }
    \end{center}
 \end{table*}

To demonstrate the ability of discovering new object classes using RGBD sensing, we experiment on a publicly available RGBD dataset~\cite{Silberman2012}. We first train a semantic segmentation network using only a subset of object classes. We then apply the proposed method to discover both the trained clases and unseen classes. We demonstrate the effectiveness of the proposed method by measuring accuracy, processing time, and memory footprint on a test dataset. All accuracy evaluations are performed at 320 $\times$ 240 resolution. Processing time is measured using a machine with an Intel Core i7-5557U 3.1GHz CPU, GeForce GTX 1080 GPU, and 16GB RAM. We use the following thresholds and constant for all the experiments: $ \sigma_{\Lambda} = 7.0, \sigma_{\Phi} = 0.8, \eta = 6.0, \alpha = 110.0, \beta = 0.5$.

\vspace{1mm}
\noindent\textbf{Dataset.}
We experiment our system using the publicly available NYUDv2 dataset~\cite{Silberman2012} which consists of 206 test video sequences. Since many of the videos have significant drops in frame-rate, they are inappropriate for tracking and reconstruction. Accordingly, previous works~\cite{Hermans2014,McCormac2017, Nakajima2018} have used only 140 test sequences that have at least 2 frames per second. This results in 360 labeled test images from the 654 images in the original test set.
%The dataset contains 206 test set video sequences, however, we picked up 140 test sequences which have a frame-rate over 2Hz as \cite{} did.

\vspace{1mm}
\noindent\textbf{U-Net Training.}
%\label{sec:traincnn}
To evaluate the proposed system's ability of class discovering, we train the U-Net using a subset of classes and evaluate the system using entire classes. This enables the quantitative analysis of both trained classes and unseen classes. We train the U-Net using the SUN RGBD training dataset~\cite{Song2015} which consists of 5,285 RGBD images. We first initialize the weights of the U-Net using the VGG model~\cite{Simonyan2015} trained on the ILSVRC dataset~\cite{Russakovsky2015}. We then finetune the model using pre-selected 9 classes among the 13 classes defined in~\cite{Couprie2013}. The selected classes and the entire classes are shown in~\tref{tab:iou}. The same trained model is used for both the proposed method and comparing methods~\cite{Ronneberger2015, Nakajima2018} in~\sref{sec:result}. 

%\subsection{Accuracy}
\subsection{Results} \label{sec:result}
We experimentally demonstrate the performance of the proposed method quantitatively and qualitatively. For quantitative comparison, we measure the Intersection over Union (IoU) using the test set of the NYUDv2 dataset~\cite{Silberman2012} and present on~\tref{tab:iou} and~\tref{tab:ablation}. In~\tref{tab:iou}, we compare the proposed method with two fully supervised methods and our methods with a different incremental 3D segmentation method~\cite{Tateno2015}. For the supervised methods, we selected one state-of-the-art semantic mapping method~\cite{Nakajima2018} and one semantic segmentation method~\cite{Ronneberger2015} for 2D images. Obviously, these methods can only predict for the 9 classes in the training dataset. 
As we propose a novel method for building a geometric 3D map using an RGBD SLIC-based segmentation method, we compare the proposed method with the method with the previous incremental 3D segmentation method of~\cite{Tateno2015}. Since~\cite{Tateno2015} uses only depth maps excluding color information, our method outperforms largely for the classes with poor geometric characteristics (\eg, picture and window). Hence, it verifies the effectiveness of the proposed SLIC-based incremental segmentation approach. Overall, the proposed method achieves competitive accuracy comparing to the state-of-the-art supervised method~\cite{Nakajima2018} and is able to successfully discover novel categories for unseen objects. Also, the proposed method outperforms the method with~\cite{Tateno2015} by 4.46 in mean IoU. 

\begin{figure}[t]
    \begin{center}
        \includegraphics[width=1.0\hsize]{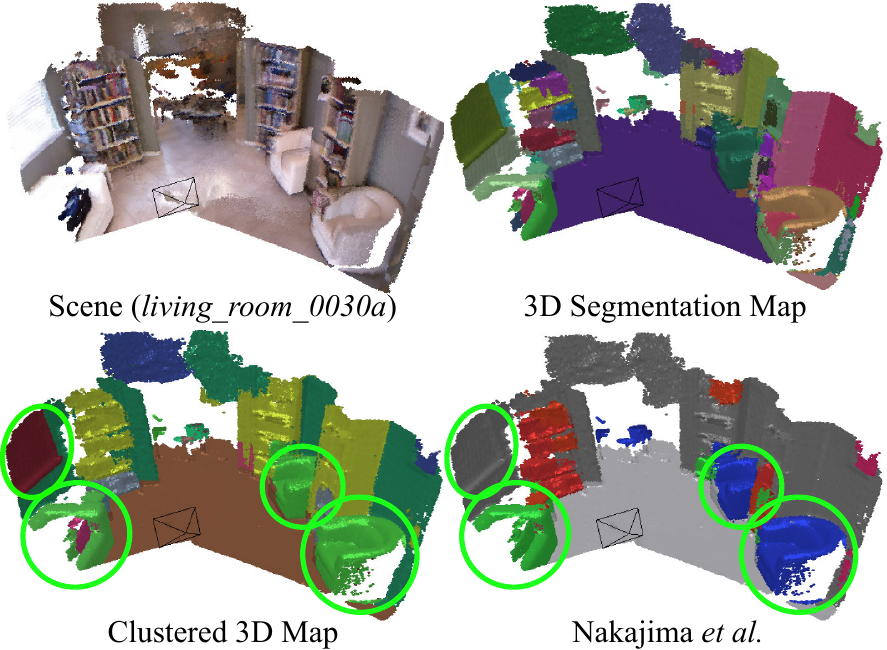}
        \caption{Qualitative results of dense 3D incremental semantic mapping. The proposed method discovers various classes including both unseen classes and the classes in the training dataset of the U-Net.
        %The U-Net is trained excluding the 4 classes (ceiling, picture, TV, and window) among 13 classes for both the proposed method and Nakajima \etal~\cite{Nakajima2018}.
        For the geometric 3D map and the clustered 3D map, a distinctive color is used for each segment and each cluster, respectively. For the results of Nakajima \etal~\cite{ Nakajima2018}, which is a fully supervised method, a specific color is used for each category as shown in~\tref{tab:iou}. \label{fig:res3d}}
%        \caption{Qualitative results of our dense 3D incremental class discovering on two scenes under the setting where the U-Net is trained without the following 4 of 13 classes, ceiling, picture, TV, and window. For the geometric 3D map and the clustered 3D map, different colors show different segments and different clusters, respectively. For the results of Nakajima et al. \cite{ Nakajima2018}, which is the fully supervised approach of semantic mapping, see Table I for class colors. \label{fig:res3d}}
    \end{center}
\end{figure}

\begin{figure}[t]
    \begin{center}
        \includegraphics[width=1.0\hsize]{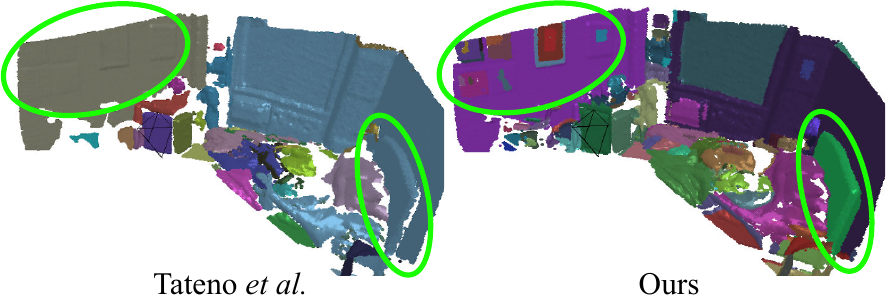}
        \caption{Qualitative results of the 3D segmentation map. The proposed method successfully segments pictures and the headboard of a bed which have poor geometric characteristics while \cite{Tateno2015} has limitation.\label{fig:res3d2}}
    \end{center}
\end{figure}

%In this section, we experimentally demonstrate the accuracy of our method with the IoU measurement through Table \ref{tab:iou}.
%In the Table \ref{tab:iou}, we include the results of the method of Nakajima et al. \cite{Nakajima2018}, which is the state-of-the-art fully supervised semantic mapping approach, with the same model of the U-Net explained in the section \ref{sec:traincnn}.
%In order to show the effectiveness of our strategy which hybridly use both of geometric and deep features, we compare against the results of ``Ours GEO-only'' and ``Ours CNN-only'', where we compute the similarity of each region $s(i,j)$ only with the geometric features $ \bm{f}^{\text{GEO}}_{l_i} $ and the deep features $ \bm{f}^{\text{CNN}}_{l_i} $ in the equation (\ref{eq:sim}), respectively.

%Table \ref{tab:iou} also shows the effectiveness of our SLIC based incremental segmentation approach by comparing with ``Ours with Tateno et al. \cite{Tateno2015}'' where we build the geometric 3D map with the method of Tateno et al. \cite{Tateno2015} that doesn't utilize the color cues.
%As shown there, the accuracy of the classes which have poor geometric characteristics (\eg, picture and window) has decreased.

In~\tref{tab:ablation}, we compare the results of the proposed method to those using only geometric features (Ours GEO-only) and those using only deep features (Ours CNN-only) to demonstrate the effectiveness of properly utilizing both features for measuring the similarity in (\ref{eq:sim}).
By comparing ``Ours GEO-only'' and ``Ours CNN-only'', we can observe that ``Ours CNN-only'' achieves higher or similar accuracy comparing to ``Ours GEO-only'' in the trained classes and ``Ours GEO-only'' outperforms ``Ours CNN-only'' for all the unseen classes. It consequently demonstrates the importance of effectively utilizing both CNN features and geometric features to achieve high accuracy in both trained classes and unseen classes. 
By applying the proposed confidence estimation, the proposed method achieves higher accuracy comparing to ``Ours GEO-only'' and ``Ours CNN-only'' in most of the classes. It verifies the effectiveness of weighting deep features and geometric features based on the estimated confidence using the entropy. The proposed method achieves 3.81 and 3.83 higher mean IoU comparing to ``Ours GEO-only'' and ``Ours CNN-only'', respectively.

%. ``GEO-only'' and ``CNN-only'' use only their corresponding features ($\bm{f}^{\text{GEO}}$ or $\bm{f}^{\text{CNN}}$) in (\ref{eq:sim}).
%which measures the similarity between segments for clustering. 

\begin{figure}[t]
    \begin{center}
        \includegraphics[width=1.0\hsize]{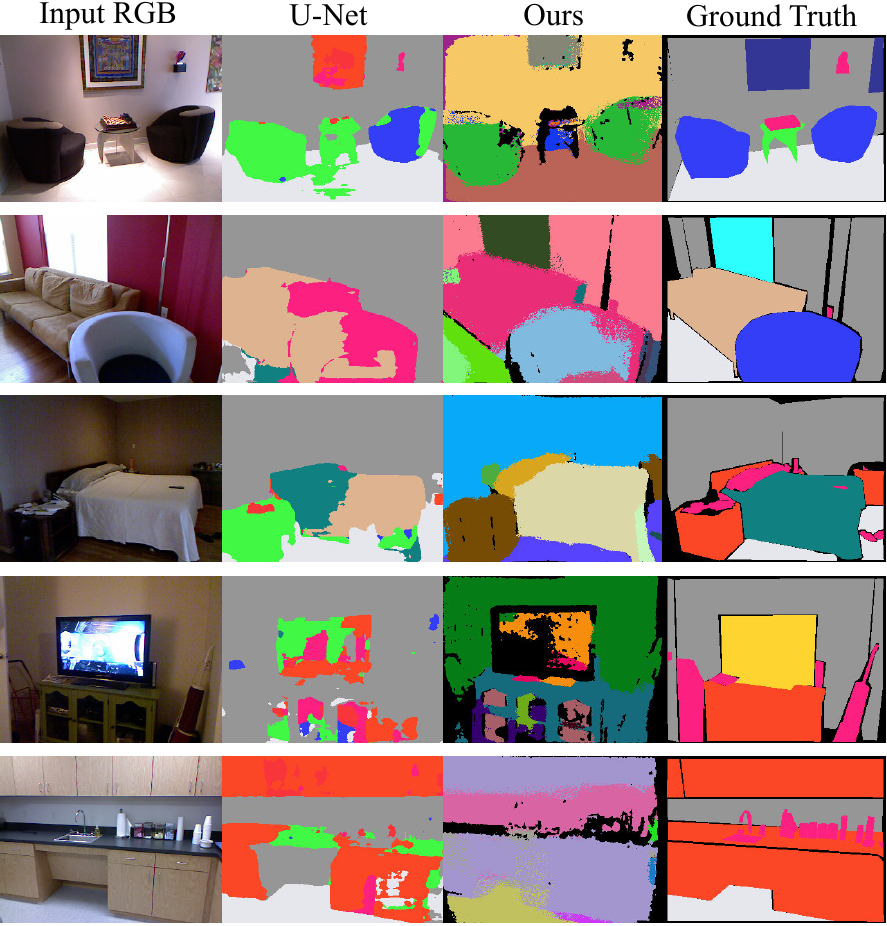}
        \caption{Qualitative comparison on the NYUDv2 dataset~\cite{Silberman2012}.
        %The U-Net is trained excluding the 4 classes (ceiling, picture, TV, and window) among 13 classes as explained in~\sref{sec:traincnn}.
        To visualize the results of the proposed method, we use a different color for each cluster. The results of the U-Net and the ground truth labels are visualized using a specific color for each category as shown in~\tref{tab:iou}. \label{fig:res2d}}
        %\caption{Qualitative comparison on the NYUDv2 dataset~\cite{Silberman2012} under the semi-supervised setting where the U-Net is trained without the following 4 of 13 classes, ceiling, picture, TV, and window. For the results of ours, different colors show different clusters. For the results of U-Net and ground truth, see Table I for class colors. \label{fig:res2d}}
    \end{center}
    \end{figure}

%As shown in Table \ref{tab:iou}, in the class categories that are not included in the training data set, \ie., unknown classes, ``Ours CNN-only'' results in less accuracy, \eg, 22.60 in picture and 42.30 in window, compared to ``Ours GEO-only'' which achieved 25.38 in picture and 52.43 in window.
%On the contrary, in the results for the class categories that are included in the training data set, \ie., known classes, ``Ours CNN-only'' achieves higher accuracy than ``Ours GEO-only''.
%For example, ``Ours CNN-only'' marks 60.07 in bed and 37.55 in chair, while ``Ours GEO-only'' marks 51.95 in bed and 35.99 in chair.
%On the other hand, the proposed method achieved the equivalent or higher accuracy than ``Ours CNN-only'' for known classes (\eg, 64.22 in ``bed'' and 41.79 in ``chair''), while achieving the equivalent or higher accuracy than ``Ours GEO-only'' for unknown classes (\eg, 28.69 in ``picture'' and 53.92 in ``window'').
%This is due to the strategy which weights the degree to which the deep features and the geometric features are utilized for measuring similarity using the entropy shown in the equation (\ref{eq:sim}).
%As a result in the mean IoU, we achieved 3.81 and 3.83 higher than the method of only using geometric features and deep features, respectively.

Figures~\ref{fig:teaser},~\ref{fig:res3d},~\ref{fig:res3d2}, and~\ref{fig:res2d} show qualitative results of the proposed method and comparing methods. The figures demonstrate that the proposed method properly clusters objects of both trained classes (for U-Net) and unseen classes. Distinctive trained objects include the chair in~\fref{fig:res3d} and the desk on~\fref{fig:teaser}. Characteristic unseen objects include the window in~\fref{fig:res3d} and the pictures in~\fref{fig:teaser}. Moreover, \fref{fig:res3d2} shows the comparison between the proposed method and~\cite{Tateno2015} in building 3D segmentation map for object proposal generation. It shows that the proposed method can segment the regions even with poor geometric characteristics (\eg, pictures on the wall) by utilizing both depth and color cues while~\cite{Tateno2015} has limitations.

%Moreover, \fref{fig:res3d} and the bottom row of~\fref{fig:teaser} show the results of the proposed object proposal generation. They demonstrate that the proposed method is able to segment even the regions with poor geometric characteristics by utilizing both depth and color cues (\eg, pictures on the wall in~\fref{fig:teaser}).

%Additionally, \fref{fig:res2d} and \fref{fig:res3d} show qualitative results of our dense 3D incremental class discovering.
%As qualitatively shown in the chair in the right scene and the desk on the left scene, the proposed method could make clusters for known objects, while making clusters even for unknown objects, \eg, pictures in the left scene and window in the right scene.
%Moreover, Fig. \ref{fig:res2d} shows the results of our object proposal approach by demonstrating that regions having poor geometric charasteristics (\eg, pictures and window on the wall in the geometric 3d map) are segmented with both of depth and color cues．
%Table \ref{tab:iou} also shows the effectiveness of our SLIC based incremental segmentation approach by comparing with ``Ours with Tateno et al. \cite{Tateno2015}'' where we build the geometric 3D map with the method of Tateno et al. \cite{Tateno2015} that doesn't utilize the color cues.
%As shown there, the accuracy of the classes which have poor geometric characteristics (\eg, picture and window) has decreased.

The bottom two rows of~\fref{fig:res2d} show the failure cases of the proposed method. On the fourth row, while the proposed method successfully segments and makes a cluster for the TV (unseen object), the furniture under the TV is segmented and grouped into multiple clusters because of the glasses on the furniture. On the fifth row, the small objects on the countertop are not segmented accurately. These kinds of objects are challenging since they are distant from the depth sensor and are small size, which often leads to less accurate depth sensing.

%The bottom two rows of \fref{fig:res2d} show failure cases of our method.
%In the fourth row of \fref{fig:res2d}, although the proposed method could successfully segment and made a class for the unknown object TV, the windows in the furniture are segmented and grouped into clusters and thus resulting in low accuracy compared to the ground truth.
%In the fifth row of \fref{fig:res2d}，since our method utilizes the vertex and normal map obtained from the incoming depth frame, it is difficult to successfully segment distant objects where depth values tend to be unstable and manage scenes where many small objects are lined up where vertices and normals are cluttered.
%We leave the exploration of improving these limitation to future work.

\subsection{Run-time Performance and Memory Footprint}
We demonstrate the efficiency of the proposed method by measuring processing time and memory footprint. The average processing time for each stage is shown in~\tref{tab:ana}. The total processing time is 93.2 ms (10.7Hz) on average. By the strategy of clustering segments instead of elements, we were able to effectively reduce the processing time of 3D segment clustering to 13.4 ms on average. The average number of segments in a 3D map was $ 253.7 $. The two most expensive processing are forward-processing of the U-Net and the feature updating. 

\begin{table}[tb]
\begin{center}
\caption{
Average processing time for each stage. Note that the processing with * and that with ** can be processed simultaneously. \label{tab:ana}
}
\scalebox{0.9}{
\begin{tabular}{lr} \bhline{1pt}
  Component & Processing time \\ \hline
  Building 3D segmentation map * & 18.2 ms \\ 
  Deep feature extraction ** & 35.9 ms \\ 
  Geometric feature extraction  & 8.2 ms \\ 
  Entropy computation  & 2.3 ms \\
  Feature/Entropy update & 33.4 ms \\
  3D segment clustering & 13.4 ms \\ \hline
  Total & {\bf 93.2 ms} \\ \bhline{1pt}
\end{tabular}
}
\end{center}
\end{table}

\begin{figure}[t]
    \begin{center}
        \includegraphics[width=1.0\hsize]{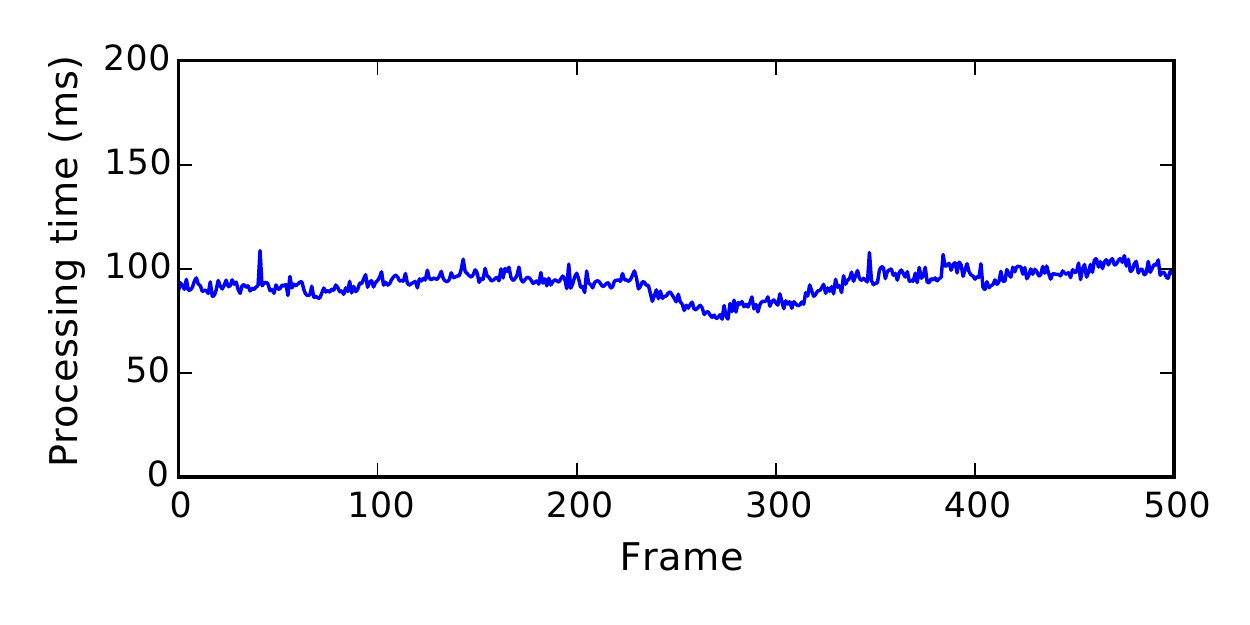}
        \caption{Processing time for each frame of the sequence {\it bedroom\_0018b} in the NYUDv2 dataset~\cite{Silberman2012}. \label{fig:time}}
    \end{center}
\end{figure}

\begin{figure}[t]
\begin{center}
    \includegraphics[width=1.0\hsize]{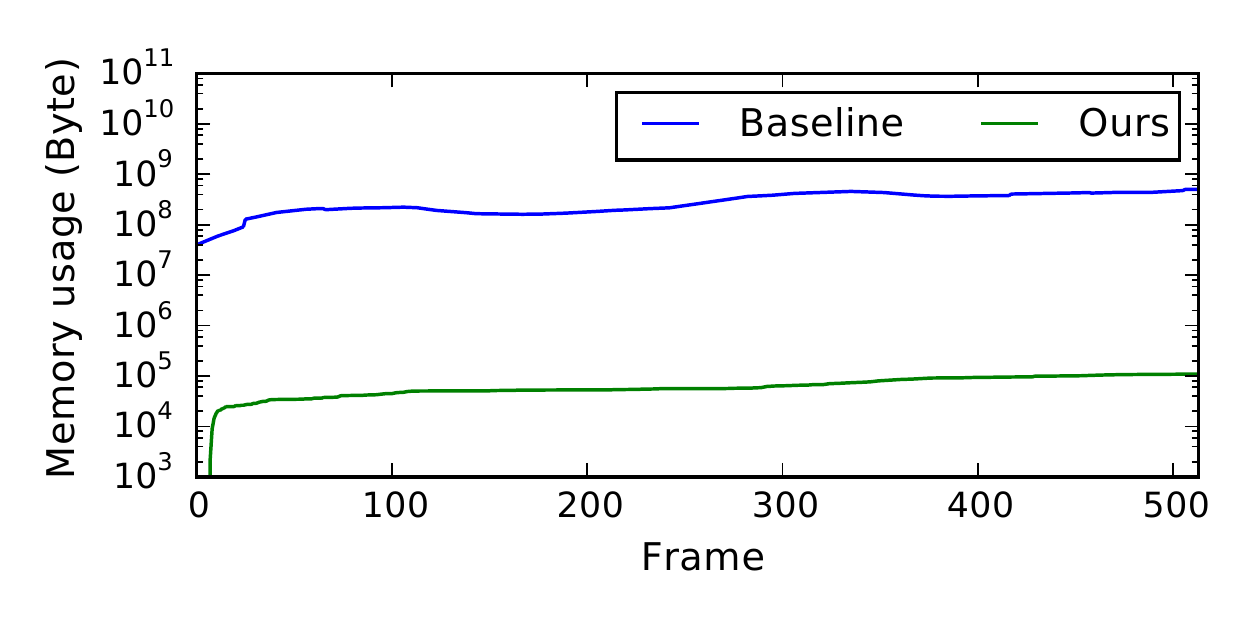}
    \caption{
    Comparison of memory usage for storing semantic and geometric features using the sequence {\it bedroom\_0018b} in the NYUDv2 dataset~\cite{Silberman2012}. While the proposed method assigns/updates features to each segment of the 3D map, the baseline method assigns/updates features to each element similar to~\cite{Hermans2014, McCormac2017} which assign class probabilities to each element. \label{fig:space}
    %Comparison of memory usage for storing semantic and geometric features. The baseline assigns features to each element of the 3D map as conventional methods assigned class probabilities to each element of the 3D map \cite{Hermans2014,McCormac2017}. This result is measured on the sequence {\it bedroom\_0018b} of the NYUDv2 dataset \cite{Silberman2012}. \label{fig:space}
    }
\end{center}
\end{figure}

We also present the processing time on~\fref{fig:time} and the memory footprint on~\fref{fig:space} for each frame in a sequence. \fref{fig:time} shows that the processing time is quite stable even though the reconstructed 3D map increases. \fref{fig:space} shows the memory footprint for storing deep features and geometric features. We compare the proposed method with the baseline method which assigns/updates features to each element similar to~\cite{Hermans2014,McCormac2017}. The analysis verifies that storing features for each segment significantly suppressed memory usage comparing to storing feature for each element. As shown in~\sref{sec:fext}, the space complexity of the proposed method is $O(N_l \cdot (S+G+1))$ while that of the baseline method is $O(N_s \cdot (S+G+1))$. After reconstructing all the frames in the sequence {\it bedroom\_0018b}, $N_l$ and $N_s$ are 196 and 900,478, respectively.

%\fref{fig:time} shows the processing time spent on each frame.
%As shown, the proposed method demonstrates an almost constant complexity even if the size of the 3D map reconstructed with SLAM is increased.
%This is because the processing for incremental segmentation (section \ref{sec:inseg}), frame-wise feature extraction(section \ref{sec:fext}), and features updates(section \ref{sec:fupdates}) limit the processing target to the incoming RGBD frames.

%Last, we discuss the results of the memory footprint used for storing semantic and geometric features through Figure \ref{fig:space}.
%We compared the proposed method with the baseline, where the features are assigned to each element of the 3D map, \ie. surfels, as conventional methods of semantic mapping assign class probabilities to each element of the 3D map \cite{Hermans2014,McCormac2017}.
%As shown there, the memory usage of the proposed method is significantly suppressed compared to the baseline over all frames.
%As analyzed in the section \ref{sec:fupdates}, the space complexity of the proposed method is $O(N_l \cdot (S+G+1))$ in contrast to the baseline takes $O(N_s \cdot (S+G+1))$, where $N_l$ and $N_s$ were 196 and 900478 in the end of the sequence {\it bedroom\_0018b}, respectively.

\section{Conclusion}

Towards open world semantic segmentation, we present a novel method that incrementally discovers new classes using RGBD sensing. We propose to discover new object classes by building a segmented dense 3D map and by identifying coherent regions in the 3D map. We demonstrate that the proposed method is able to successfully discover new object classes by experimenting on a public dataset. The experimental results also show that the proposed method achieves competitive accuracy for known classes comparing to the supervised methods. We further show that the proposed method is very efficient in computation and memory usage.

{\small
\bibliographystyle{ieee}
\bibliography{egbib}

\begin{thebibliography}{10}\itemsep=-1pt

\bibitem{Arbelaez2009}
P.~{Arbelaez}, M.~{Maire}, C.~{Fowlkes}, and J.~{Malik}.
\newblock From contours to regions: An empirical evaluation.
\newblock In {\em 2009 IEEE Conference on Computer Vision and Pattern
  Recognition}, pages 2294--2301, June 2009.

\bibitem{Arbelaez2011}
P.~{Arbelaez}, M.~{Maire}, C.~{Fowlkes}, and J.~{Malik}.
\newblock Contour detection and hierarchical image segmentation.
\newblock {\em IEEE Transactions on Pattern Analysis and Machine Intelligence},
  33(5):898--916, May 2011.

\bibitem{Boykov2001}
Y.~{Boykov}, O.~{Veksler}, and R.~{Zabih}.
\newblock Fast approximate energy minimization via graph cuts.
\newblock {\em IEEE Transactions on Pattern Analysis and Machine Intelligence},
  23(11):1222--1239, Nov 2001.

\bibitem{Chen2018}
L.~{Chen}, G.~{Papandreou}, I.~{Kokkinos}, K.~{Murphy}, and A.~L. {Yuille}.
\newblock Deeplab: Semantic image segmentation with deep convolutional nets,
  atrous convolution, and fully connected crfs.
\newblock {\em IEEE Transactions on Pattern Analysis and Machine Intelligence},
  40(4):834--848, April 2018.

\bibitem{Comaniciu2002}
D.~{Comaniciu} and P.~{Meer}.
\newblock Mean shift: a robust approach toward feature space analysis.
\newblock {\em IEEE Transactions on Pattern Analysis and Machine Intelligence},
  24(5):603--619, May 2002.

\bibitem{Couprie2013}
C.~Couprie, C.~Farabet, L.~Najman, and Y.~LeCun.
\newblock Indoor semantic segmentation using depth information.
\newblock In {\em International Conference on Learning Representations}, 2013.

\bibitem{Deng2001}
Y.~{Deng} and B.~S. {Manjunath}.
\newblock Unsupervised segmentation of color-texture regions in images and
  video.
\newblock {\em IEEE Transactions on Pattern Analysis and Machine Intelligence},
  23(8):800--810, Aug 2001.

\bibitem{Felzenszwalb2004}
P.~F. Felzenszwalb and D.~P. Huttenlocher.
\newblock Efficient graph-based image segmentation.
\newblock {\em International Journal of Computer Vision}, 59(2):167--181, Sep
  2004.

\bibitem{Fulkerson2012}
B.~Fulkerson and S.~Soatto.
\newblock Really quick shift: Image segmentation on a gpu.
\newblock In K.~N. Kutulakos, editor, {\em Trends and Topics in Computer
  Vision}, pages 350--358, Berlin, Heidelberg, 2012. Springer Berlin
  Heidelberg.

\bibitem{Hermans2014}
A.~{Hermans}, G.~{Floros}, and B.~{Leibe}.
\newblock Dense 3d semantic mapping of indoor scenes from rgb-d images.
\newblock In {\em 2014 IEEE International Conference on Robotics and Automation
  (ICRA)}, pages 2631--2638, May 2014.

\bibitem{Huang2004}
Y.-L. Huang and D.-R. Chen.
\newblock Watershed segmentation for breast tumor in 2-d sonography.
\newblock {\em Ultrasound in Medicine and Biology}, 30(5):625 -- 632, 2004.

\bibitem{Izadi2011}
S.~Izadi, D.~Kim, O.~Hilliges, D.~Molyneaux, R.~Newcombe, P.~Kohli, J.~Shotton,
  S.~Hodges, D.~Freeman, A.~Davison, and A.~Fitzgibbon.
\newblock Kinectfusion: Real-time 3d reconstruction and interaction using a
  moving depth camera.
\newblock In {\em Proceedings of the 24th Annual ACM Symposium on User
  Interface Software and Technology}, UIST '11, pages 559--568, New York, NY,
  USA, 2011. ACM.

\bibitem{kasaei2016good}
S.~H. Kasaei, A.~M. Tom{\'e}, L.~S. Lopes, and M.~Oliveira.
\newblock Good: A global orthographic object descriptor for 3d object
  recognition and manipulation.
\newblock {\em Pattern Recognition Letters}, 83:312--320, 2016.

\bibitem{Keller2013}
M.~{Keller}, D.~{Lefloch}, M.~{Lambers}, S.~{Izadi}, T.~{Weyrich}, and
  A.~{Kolb}.
\newblock Real-time 3d reconstruction in dynamic scenes using point-based
  fusion.
\newblock In {\em 2013 International Conference on 3D Vision - 3DV 2013}, pages
  1--8, June 2013.

\bibitem{Koppula2011}
H.~S. Koppula, A.~Anand, T.~Joachims, and A.~Saxena.
\newblock Semantic labeling of 3d point clouds for indoor scenes.
\newblock In J.~Shawe-Taylor, R.~S. Zemel, P.~L. Bartlett, F.~Pereira, and
  K.~Q. Weinberger, editors, {\em Advances in Neural Information Processing
  Systems 24}, pages 244--252. Curran Associates, Inc., 2011.

\bibitem{Kundu2014}
A.~Kundu, Y.~Li, F.~Dellaert, F.~Li, and J.~M. Rehg.
\newblock Joint semantic segmentation and 3d reconstruction from monocular
  video.
\newblock In D.~Fleet, T.~Pajdla, B.~Schiele, and T.~Tuytelaars, editors, {\em
  Computer Vision -- ECCV 2014}, pages 703--718, Cham, 2014. Springer
  International Publishing.

\bibitem{Lee2016}
K.-R. Lee and T.~Nguyen.
\newblock Realistic surface geometry reconstruction using a hand-held rgb-d
  camera.
\newblock {\em Machine Vision and Applications}, 27(3):377--385, Apr 2016.

\bibitem{Li2017}
X.~{Li}, H.~{Ao}, R.~{Belaroussi}, and D.~{Gruyer}.
\newblock Fast semi-dense 3d semantic mapping with monocular visual slam.
\newblock In {\em 2017 IEEE 20th International Conference on Intelligent
  Transportation Systems (ITSC)}, pages 385--390, Oct 2017.

\bibitem{Long2015}
J.~{Long}, E.~{Shelhamer}, and T.~{Darrell}.
\newblock Fully convolutional networks for semantic segmentation.
\newblock In {\em 2015 IEEE Conference on Computer Vision and Pattern
  Recognition (CVPR)}, pages 3431--3440, June 2015.

\bibitem{McCormac2017}
J.~{McCormac}, A.~{Handa}, A.~{Davison}, and S.~{Leutenegger}.
\newblock Semanticfusion: Dense 3d semantic mapping with convolutional neural
  networks.
\newblock In {\em 2017 IEEE International Conference on Robotics and Automation
  (ICRA)}, pages 4628--4635, May 2017.

\bibitem{Nakajima2018}
Y.~{Nakajima}, K.~{Tateno}, F.~{Tombari}, and H.~{Saito}.
\newblock Fast and accurate semantic mapping through geometric-based
  incremental segmentation.
\newblock In {\em 2018 IEEE/RSJ International Conference on Intelligent Robots
  and Systems (IROS)}, pages 385--392, Oct 2018.

\bibitem{Newcombe2011}
R.~A. {Newcombe}, S.~{Izadi}, O.~{Hilliges}, D.~{Molyneaux}, D.~{Kim}, A.~J.
  {Davison}, P.~{Kohi}, J.~{Shotton}, S.~{Hodges}, and A.~{Fitzgibbon}.
\newblock Kinectfusion: Real-time dense surface mapping and tracking.
\newblock In {\em 2011 10th IEEE International Symposium on Mixed and Augmented
  Reality}, pages 127--136, Oct 2011.

\bibitem{nguyen2012modeling}
C.~V. Nguyen, S.~Izadi, and D.~Lovell.
\newblock Modeling kinect sensor noise for improved 3d reconstruction and
  tracking.
\newblock In {\em 2012 second international conference on 3D imaging, modeling,
  processing, visualization \& transmission}, pages 524--530. IEEE, 2012.

\bibitem{Pont2017}
J.~{Pont-Tuset}, P.~{Arbeláez}, J.~{T. Barron}, F.~{Marques}, and J.~{Malik}.
\newblock Multiscale combinatorial grouping for image segmentation and object
  proposal generation.
\newblock {\em IEEE Transactions on Pattern Analysis and Machine Intelligence},
  39(1):128--140, Jan 2017.

\bibitem{Prisacariu2017}
V.~A. Prisacariu, O.~K{\"{a}}hler, S.~Golodetz, M.~Sapienza, T.~Cavallari,
  P.~H.~S. Torr, and D.~W. Murray.
\newblock Infinitam v3: {A} framework for large-scale 3d reconstruction with
  loop closure.
\newblock {\em CoRR}, abs/1708.00783, 2017.

\bibitem{Ray1999}
S.~Ray and R.~H. Turi.
\newblock Determination of number of clusters in k-means clustering and
  application in colour segmentation.
\newblock In {\em The 4th International Conference on Advances in Pattern
  Recognition and Digital Techniques}, pages 137--143, 1999.

\bibitem{Ronneberger2015}
O.~Ronneberger, P.~Fischer, and T.~Brox.
\newblock U-net: Convolutional networks for biomedical image segmentation.
\newblock In N.~Navab, J.~Hornegger, W.~M. Wells, and A.~F. Frangi, editors,
  {\em Medical Image Computing and Computer-Assisted Intervention -- MICCAI
  2015}, pages 234--241, Cham, 2015. Springer International Publishing.

\bibitem{Russakovsky2015}
O.~Russakovsky, J.~Deng, H.~Su, J.~Krause, S.~Satheesh, S.~Ma, Z.~Huang,
  A.~Karpathy, A.~Khosla, M.~Bernstein, A.~C. Berg, and L.~Fei-Fei.
\newblock Imagenet large scale visual recognition challenge.
\newblock {\em International Journal of Computer Vision}, 115(3):211--252, Dec
  2015.

\bibitem{Sengupta2013}
S.~{Sengupta}, E.~{Greveson}, A.~{Shahrokni}, and P.~H.~S. {Torr}.
\newblock Urban 3d semantic modelling using stereo vision.
\newblock In {\em 2013 IEEE International Conference on Robotics and
  Automation}, pages 580--585, May 2013.

\bibitem{Shannon1948}
C.~E. Shannon.
\newblock A mathematical theory of communication.
\newblock {\em Bell System Technical Journal}, 27(3):379--423, 1948.

\bibitem{Shelhamer2017}
E.~{Shelhamer}, J.~{Long}, and T.~{Darrell}.
\newblock Fully convolutional networks for semantic segmentation.
\newblock {\em IEEE Transactions on Pattern Analysis and Machine Intelligence},
  39(4):640--651, April 2017.

\bibitem{Shi2000}
J.~Shi and J.~{Malik}.
\newblock Normalized cuts and image segmentation.
\newblock {\em IEEE Transactions on Pattern Analysis and Machine Intelligence},
  22(8):888--905, Aug 2000.

\bibitem{Silberman2012}
N.~Silberman, D.~Hoiem, P.~Kohli, and R.~Fergus.
\newblock Indoor segmentation and support inference from rgbd images.
\newblock In A.~Fitzgibbon, S.~Lazebnik, P.~Perona, Y.~Sato, and C.~Schmid,
  editors, {\em Computer Vision -- ECCV 2012}, pages 746--760, Berlin,
  Heidelberg, 2012. Springer Berlin Heidelberg.

\bibitem{Simonyan2015}
K.~Simonyan and A.~Zisserman.
\newblock Very deep convolutional networks for large-scale image recognition.
\newblock In {\em International Conference on Learning Representations}, 2015.

\bibitem{Song2015}
S.~{Song}, S.~P. {Lichtenberg}, and J.~{Xiao}.
\newblock Sun rgb-d: A rgb-d scene understanding benchmark suite.
\newblock In {\em 2015 IEEE Conference on Computer Vision and Pattern
  Recognition (CVPR)}, pages 567--576, June 2015.

\bibitem{Tateno2015}
K.~{Tateno}, F.~{Tombari}, and N.~{Navab}.
\newblock Real-time and scalable incremental segmentation on dense slam.
\newblock In {\em 2015 IEEE/RSJ International Conference on Intelligent Robots
  and Systems (IROS)}, pages 4465--4472, Sep. 2015.

\bibitem{VanDongen2000}
S.~van Dongen.
\newblock {Graph clustering by flow simulation}.
\newblock {\em University of Utrecht}, 2000.

\bibitem{Vineet2015}
V.~{Vineet}, O.~{Miksik}, M.~{Lidegaard}, M.~{Nießner}, S.~{Golodetz}, V.~A.
  {Prisacariu}, O.~{Kähler}, D.~W. {Murray}, S.~{Izadi}, P.~{Pérez}, and
  P.~H.~S. {Torr}.
\newblock Incremental dense semantic stereo fusion for large-scale semantic
  scene reconstruction.
\newblock In {\em 2015 IEEE International Conference on Robotics and Automation
  (ICRA)}, pages 75--82, May 2015.

\bibitem{Xia2017}
X.~Xia and B.~Kulis.
\newblock W-net: {A} deep model for fully unsupervised image segmentation.
\newblock {\em CoRR}, abs/1711.08506, 2017.

\bibitem{Yang2015}
J.~{Yang}, Z.~{Gan}, K.~{Li}, and C.~{Hou}.
\newblock Graph-based segmentation for rgb-d data using 3-d geometry enhanced
  superpixels.
\newblock {\em IEEE Transactions on Cybernetics}, 45(5):927--940, May 2015.

\bibitem{Yang2017}
S.~{Yang}, Y.~{Huang}, and S.~{Scherer}.
\newblock Semantic 3d occupancy mapping through efficient high order crfs.
\newblock In {\em 2017 IEEE/RSJ International Conference on Intelligent Robots
  and Systems (IROS)}, pages 590--597, Sep. 2017.

\bibitem{Zhang2012}
Z.~{Zhang}.
\newblock Microsoft kinect sensor and its effect.
\newblock {\em IEEE MultiMedia}, 19(2):4--10, Feb 2012.

\end{thebibliography}
}

\end{document}